\documentclass[letterpaper, 10 pt, conference]{ieeeconf}  

\IEEEoverridecommandlockouts                              

\title{\LARGE \bf
Cloud-Edge Training Architecture for \\ Sim-to-Real Deep Reinforcement Learning
}

\author{Hongpeng Cao, Mirco Theile, Federico G. Wyrwal, and Marco Caccamo
\thanks{Hongpeng Cao, Mirco Theile, Federico G. Wyrwal, and Marco Caccamo are  with the Technical University of Munich (TUM), School of Engineering and Design, Munich, Germany
{\tt\small \{cao.hongpeng, mirco.theile, federico.wyrwal, mcaccamo\}@tum.de}}%
}

\usepackage{amsfonts}
\usepackage{amsmath}
\usepackage{amssymb}
\usepackage[]{todonotes}
\usepackage{graphicx}
\usepackage{hyperref}
\usepackage{caption}
\usepackage{subcaption}

\usepackage{tabularx}
\usepackage{booktabs}
\usepackage{balance}
\usepackage{color,soul}
\begin{document}

\maketitle
\thispagestyle{empty}
\pagestyle{empty}

\begin{abstract}
Deep reinforcement learning (DRL) is a promising approach to solve complex control tasks by learning policies through interactions with the environment. However, the training of DRL policies requires large amounts of training experiences, making it impractical to learn the policy directly on physical systems. Sim-to-real approaches leverage simulations to pretrain DRL policies and then deploy them in the real world. Unfortunately, the direct real-world deployment of pretrained policies usually suffers from performance deterioration due to the different dynamics, known as the \textit{reality gap}. Recent sim-to-real methods, such as domain randomization and domain adaptation, focus on improving the robustness of the pretrained agents. Nevertheless, the simulation-trained policies often need to be tuned with real-world data to reach optimal performance, which is challenging due to the high cost of real-world samples. 

This work proposes a distributed cloud-edge architecture to train DRL agents in the real world in real-time. In the architecture, the inference and training are assigned to the edge and cloud, separating the real-time control loop from the computationally expensive training loop. To overcome the \textit{reality gap}, our architecture exploits sim-to-real transfer strategies to continue the training of simulation-pretrained agents on a physical system. We demonstrate its applicability on a physical inverted-pendulum control system, analyzing critical parameters. The real-world experiments show that our architecture can adapt the pretrained DRL agents to unseen dynamics consistently and efficiently. \footnote{A video showing a real-world training process under the proposed method can be found from \url{https://youtu.be/hMY9-c0SST0}.}
\end{abstract}
\section{Introduction}

Deep reinforcement learning (DRL) enables robots to master complicated tasks with human-level performance. Instead of learning from labeled data, the DRL agent learns control policies via interactions with the environment, aiming to maximize cumulative rewards. Recent progress shows that DRL can achieve impressive performances in the robotic control domain, e.g., in locomotion \cite{Haarnoja18}, grasping \cite{levine2016end}, and manipulation \cite{zhu2019dexterous}.

However, due to the high demand for data for DRL, direct training on physical systems presents many challenges, as discussed in \cite{kober2013reinforcement, ibarz2021train}. Collecting training data in the real world is expensive concerning time and labor. Human supervision is usually needed to reset the system and monitor its hardware maintenance and safety status. Recent work in the field aims to address these challenges. In \cite{Haarnoja18}, an improved soft actor-critic algorithm \cite{haarnoja2018soft} is proposed to reduce the learning sensitivities to hyperparameter settings, making training on physical systems more stable. Approaches of off-policy training with replay memory \cite{Kalashnikov2018, gu2017deep} and model-based reinforcement learning \cite{nagabandi2018learning, Yang2020} can significantly increase real-world sampling efficiency. Moreover, training using demonstrations \cite{Vecerik2017} and scripted policies \cite{Johannink2019} can ease the real-world exploration. When training on hardware, additional problems are that the dynamics of the systems might be non-stationary due to hardware wear and tear, making long training harder or even impossible. Environmental noise and disturbances can further exacerbate this issue. Moreover, physical systems' inherent sensing, actuating, computation, and data communication delays often violate the Markovian assumption, a prerequisite for reinforcement learning. 

In contrast, modern simulations can simulate complex systems and various environments. Recent sim-to-real approaches pretrain agents in simulations and then directly deploy the learned policies for real-world applications without further training. Simulation-based training boosts sampling efficiency significantly since simulations run faster than physical systems and can be further improved via parallel training. Additionally, the system can reset automatically without human intervention. Simulation training also does not require hardware maintenance and safety measures, drastically reducing the need for human supervision. Unfortunately, direct deployment of simulation-trained agents often fails in real-world applications due to the dynamic divergence between the real world and the simulated environment, the so-called \textit{reality gap}.The \textit{reality gap} arises mainly from system under-modeling \cite{kober2013reinforcement, neunert2017off}, where the complex dynamics, e.g., the contact and friction effects, are difficult to measure and model. Moreover, computation and communication delays and environmental noise introduce extra modeling errors. System identification approaches such as \cite{Tan2018} aim to create more accurate simulations, mitigating modeling errors. However, most of the sim-to-real approaches utilize domain randomization \cite{Peng2018, Sadeghi2017, andrychowicz2020learning} or domain adaptation \cite{James2019, Chiang2019} to improve the agents' robustness in simulations, resulting in a successful direct real-world deployment. In many scenarios, the domain randomization strategies might not be feasible since there might not exist a single policy that can solve all problems within the domain.

Pretraining in simulation to learn a sub-optimal policy and then continuing the training in the real world can take advantage of the efficient simulation training and the real dynamics. With this approach, all the advantages of simulation learning can be adopted, and the \textit{reality gap} can be closed while training in the real world. We follow the continuous sim-to-real training paradigm in this work, as we believe it is the most promising approach for real-world DRL. However, the continuous sim-to-real training paradigm suffers from two main problems. First, DRL requires high-performance computation and benefits from dedicated devices such as GPUs or TPUs for its training loop. In many real-time control systems, e.g., unmanned aerial vehicles or other mobile robots, the plant cannot have a high-performance device onboard due to power, weight, and space constraints. It cannot be controlled directly from a remote high-performance device, as perfect, loss-less, and low-delay communication cannot be guaranteed. The second problem is that the transfer of a pretrained agent to the real physical system is non-trivial. The dynamics can change abruptly, making the value estimates of the DRL agent inaccurate, leading to deteriorating performance. 

This work addresses the real-world training problem by introducing a novel distributed cloud-edge architecture. The real-time control loop on the edge is decoupled from the computationally intensive training loop on the cloud. The agent on the edge collects experiences in real-time, sending them to the trainer on the cloud, which periodically updates the edge with the optimized parameters. The agent is double-buffered such that the real-time inference loop is not interrupted by the policy updates. We address the sim-to-real transfer problem by delaying the neural network training at the beginning of the real-world interactions. Specifically, we start the policy optimization later than the value estimate training to avoid policy deterioration based on unstable value estimates. 

We evaluate our approach on a physical inverted-pendulum system controlled by a DRL agent deployed on a Raspberry Pi 4B. The training loop is offloaded to a high-performance workstation. The agents are pretrained in intentionally under-modeled simulations to induce different levels of the \textit{reality gap} to analyze the sim-to-real transfer. We further analyze the impact of the neural network optimization delays, highlighting their necessity. Additionally, we investigate the relevance of the edge update frequency on training performance to show that the architecture can work in constrained bandwidth settings, albeit with an impact on training time. Our contributions can be summarized as follows:
\begin{itemize}
    \item Design of a distributed cloud-edge architecture that enables continuous sim-to-real training for DRL agents;
    \item Conception and evaluation of sim-to-real transfer methods that mitigate policy performance deterioration after deployment to the physical system;
    \item Evaluation and analysis of our approach using an off-policy actor-critic DRL method to control a physical inverted pendulum system with the actor deployed on an embedded system.
\end{itemize}

This work is structured as follows: Section \ref{sec:background} describes the background needed for the proposed architecture and sim-to-real transfer strategies in Section \ref{sec:method}. The proposed approach is evaluated with experimental setup discussed in Section \ref{sec:setup} and results presented in Section \ref{sec:results}. Section \ref{sec:conclusion} concludes the work and gives a brief outlook on future work.
\section{Background}
\label{sec:background}
This section introduces the basics of RL methods and describes the foundations of the Deep Deterministic Policy Gradient (DDPG) algorithm \cite{lillicrap2015continuous}, an off-policy algorithm, which we adapted for the proposed cloud-edge architecture.

\subsection{Reinforcement Learning}

A DRL agent is interacting with its environment in discrete timesteps, which can be formulated as a Markov Decision Process (MDP) with $\mathcal{M} = \{\mathcal{S}, \mathcal{A}, P, R,\gamma\}$. 
In the MDP, 
$\mathcal{S}$ represents a set of states, 
$\mathcal{A}$ a set of actions, and
$P : \mathcal{S} \times \mathcal{A} \times \mathcal{S} \mapsto \mathbb{R}$
the state-transition probability function indicating the probability of a state-action pair leading to a specific next state.
The reward function $R : \mathcal{S} \times \mathcal{A} \times \mathcal{S} \mapsto \mathbb{R}$
maps a state-action-next state triple to a real-valued reward.
The discount factor $\gamma \in [0, 1]$ controls the relative importance of immediate and future rewards. The goal in DRL is to find a policy $\pi: \mathcal{S} \mapsto \mathcal{A}$, mapping a state to an action that maximizes the expected return from step $t$ 
\begin{equation}
    G_t = \sum_{i=t}^{\infty} {\gamma^{i-t} R(s_i, a_i, s_{i+1})}.
\end{equation}
To find a policy that maximizes the return, Q-learning approaches estimate the state-action value
\begin{equation}
    Q^\pi(s,a) = \mathbb{E}_{\pi} \left[G_t | s_t = s, a_t = a\right],
    \label{eq:q}
\end{equation} 
as the return if taking action $a$ at state $s$ and following policy $\pi$ afterwards.

\subsection{Deep Deterministic Policy Gradient}
DDPG is an off-policy actor-critic algorithm, in which the actor is parameterized using a deep neural network with parameters $\theta$, creating the policy approximate $\pi_\theta$. The critic is parameterized by a deep neural network with parameters $\phi$ to estimate the state-action value $Q_\phi(s, a)$. During training, the critic network is trained to minimize the expectation of the temporal difference (TD) error 
\begin{equation}
    L(\phi) = \mathbb{E}\left[ (Q_\phi(s_t,a_t) - y_t)^2 \right],
\end{equation}
where
\begin{equation}
    y_t = R(s_t,a_t,s_{t+1}) + \gamma (1 - \beta(s_{t+1})) Q_{\bar{\phi}}(s_{t+1}, \pi_{\bar{\theta}}(s_{t+1})),
    \label{eq:q_target}
\end{equation}
with $\beta(s_{t+1})$ indicating whether $s_{t+1}$ is a terminal state. The value and action at the next state are estimated by the critic and actor target networks parameterized with $\bar{\phi}$ and $\bar{\theta}$, respectively.

The actor is aiming to maximize the value estimate of the critic and is optimized using the deterministic gradient
\begin{equation}
    \nabla_\theta Q_\phi(s_t, \pi_\theta(s_t)) = \nabla_a Q_\phi(s_t, a) \nabla_\theta \pi_\theta(s_t).
\end{equation}

DDPG also proposes to soft update the target networks by slowly tracking the learned network: $\bar{\phi} \leftarrow \tau\phi + (1-\tau)\bar{\phi}$ with $\tau \in (0, 1]$ and similarly for $\bar{\theta}$. The resulting moving-average over the network parameters stabilizes training.

Twin Delayed Deep Deterministic Policy Gradient (TD3) algorithm \cite{td3}, a variant of the DDPG family, utilizes delayed policy optimization, target policy smoothing regularization, and clipped double Q-learning. In our simulation, we found that only policy smoothing regularization helps the training converge faster, while the others do not show improvements. Thus, we only adopt the idea of policy smoothing regularization used in TD3 by replacing the $\pi_{\bar{\theta}}(s_{t+1})$ in \eqref{eq:q_target} with
\begin{equation}\label{target_action_with_noise}
    \operatorname{clip}(\pi_{\bar{\theta}}(s_{t+1})  + a_{\mathcal{N}}, -a_{\text{max}}, a_{\text{max}}),
\end{equation}
which regularizes the target action by adding a clipped action noise $a_{\mathcal{N}} = \operatorname{clip}(\epsilon, -c, c)$ where $\epsilon$ is sampled from a Gaussian distributions $\mathcal{N}(0, \sigma)$. The $\operatorname{clip}(z,z_\text{min},z_\text{max})$ function constraints the value of $z$ to the range $[z_\text{min}, z_\text{max}]$.

\vspace{-5pt}
\subsection{Combined Experience Replay}
Experience Replay (ER) is a technique to train the agent with a batch of $B$ experiences sampled from a buffer of previous transitions. ER is used in DDPG and other recent off-policy RL algorithms to provide uncorrelated data for training deep neural networks \cite{mnih2015human} and improve sampling efficiency significantly \cite{lin1993reinforcement}. 

A common method for improving replay memory efficiency is to use prioritized experience replay \cite{schaulprio}. We chose not to use prioritized replay because we hypothesize that for experiences collected in the real world, prioritization may increase the sampling probability of corrupted or abnormal experiences, which arise from noise or other disturbances. We will investigate this hypothesis in future work.

Instead, we use Combined Experience Replay (CER), proposed by Zhang et al. in \cite{zhang2017deeper}, which aims to remedy the training sensitivity to the buffer capacity with very low computational complexity $\mathcal{O}(1)$. In CER, only $B-1$ samples are sampled from memory, and the agent's latest experience is always added. This allows the training process to react quickly to newly observed transitions.
\section{Methodology}
\label{sec:method}
This section presents our continuous training method: a distributed cloud-edge training architecture and the sim-to-real transfer learning strategies.

\subsection{Remote Training Architecture}

\begin{figure}
\centering
\vspace{5pt}
\includegraphics[width=0.98 \columnwidth ]{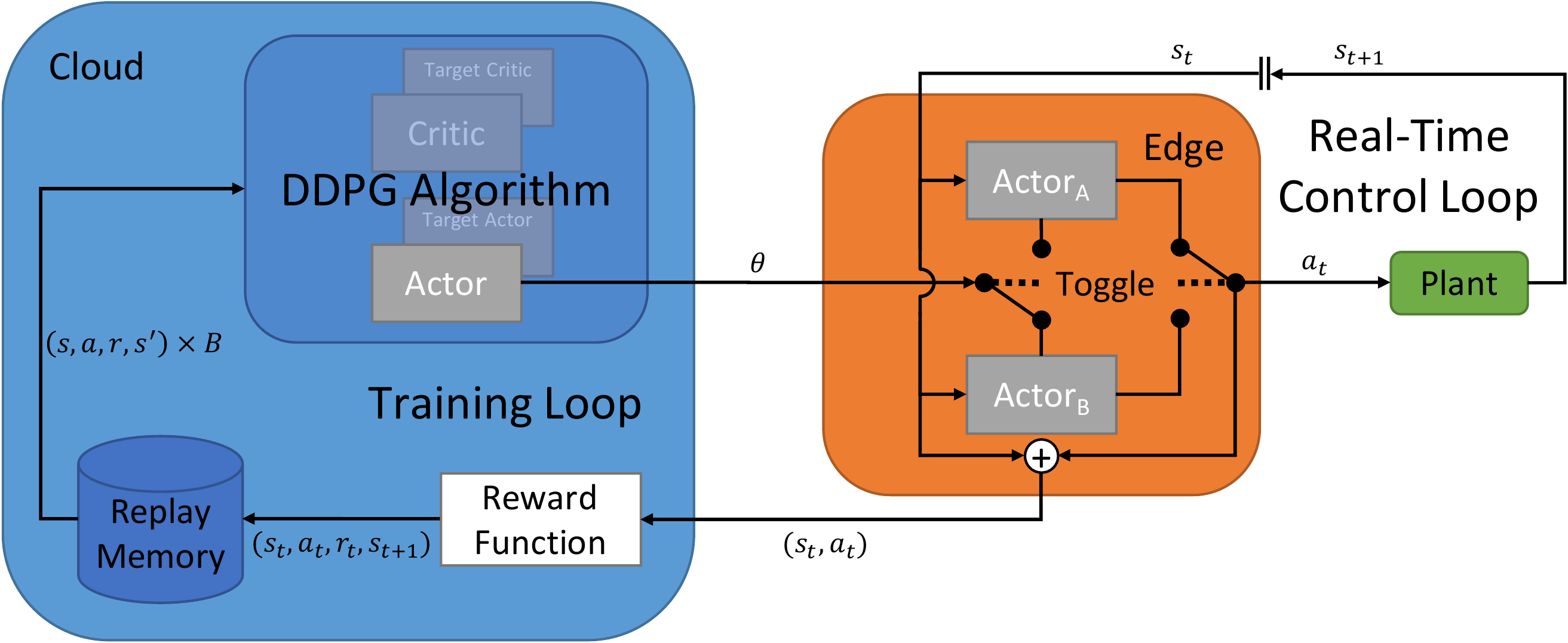}
\caption{Distributed cloud-edge training architecture displaying the cloud, edge, and plant with the cloud and edge interacting through the training loop, and the edge and plant interacting through the real-time control loop; the loops are disconnected by double-buffering the actor on the edge.}

\label{fig:diagram}
\end{figure}

We devised a distributed cloud-edge training architecture to minimize the computational load on the embedded device used to control the plant. To this end, most of the computation is offloaded to a high-performance device, the cloud, as shown in Figure \ref{fig:diagram}. The diagram shows the three devices, the cloud, the edge, and the plant. The cloud contains the reward function, replay memory, and the full actor-critic setup of the DDPG algorithm. The edge only contains the actor network. The plant represents the physical control system, including actuators and sensors, and is directly connected to the edge. The cloud and edge can be connected via any networking protocol. Two decoupled interaction loops arise by double-buffering the actor on the edge, a real-time control loop between the edge and plant, and a training loop between the cloud and edge.

\subsubsection{Real-Time Control Loop}
In the real-time control loop, the currently active actor-network on the edge ($\text{Actor}_\text{A}$ in Figure \ref{fig:diagram}) infers an action $a_t$ based on the observed state $s_t$, which is passed to the actuators of the plant. This interaction loop is real-time critical, meaning that it has to be temporally deterministic and predictable. This requirement arises from the necessity of the Markov property for reinforcement learning. In a system that fulfills the Markov property, the transition probability function $P$ only depends on the current state $s_t$ and current action $a_t$ and is independent of the past states and actions \cite{sutton2018reinforcement}. On physical control systems, this property can be easily violated by sensing, computation, actuation, and communication delays, which can make the next state depending on previous actions. This problem can be mitigated by making past actions or states part of the observation space or adding recurrence to the policy \cite{ibarz2021train}. 

However, these augmentations only create a Markovian observation under consistent or predictable delays. Consistent delays are achieved by scheduling computation and communication in fixed time slots with a fixed period. This mimics traditional real-time control systems \cite{buttazzo2011hard}. Fixed periods can only be achieved if the computation time has an upper bound smaller than the period. In the proposed architecture, the computation time is given by the inference time of the actor-network on the edge. By decoupling the inference from the training utilizing the cloud and from policy updates by double-buffering the actor, we enable its real-time capabilities. 

\subsubsection{Training Loop}
The state-action pairs $(s_t, a_t)$ created through the interaction in the control loop are sent to the cloud. The reward function $R$ uses the state $s_t$, the action $a_t$, and the next state $s_{t+1}$ to compute the reward $r_t$ for the experience tuple $(s_t, a_t, r_t, s_{t+1})$, which is added to the replay memory. The time index is removed in the replay memory, as it is not needed for off-policy training. A batch of $B$ experiences is sampled from the replay memory and passed to the DDPG algorithm. The relative next states in the batch are indicated with $s^\prime$. The algorithm updates the critics and actors, sending the new actor parameters $\theta$ to the edge. On the edge, the parameters are applied to the inactive actor ($\text{Actor}_\text{B}$ in Figure \ref{fig:diagram}), toggling the active actor on completion.

\subsection{Sim-to-Real Transfer}
As mentioned before, we augmented the observation space with the previous five actions to overcome non-Markovianess due to delays. However, the agent learned to ignore these extra observations in simulation, as they were not needed in the non-delayed simulation environment. We introduced a one-step delay in simulation, such that the extra observations cannot be ignored.

The main problem when transferring to the real world arises from the replay buffer. A certain amount of experiences is needed in the buffer, such that when sampling from it, the experience batch is uncorrelated. One potential remedy would be to keep experiences from simulation and gradually replace them with real-world data. However, this approach could be sensitive to the size of the \textit{reality gap} because the transition probability function $P$ might be significantly different. We will investigate this assumption in the future. Instead, we adopted a solution that is independent of the size of the \textit{reality gap} by starting with an empty replay memory and collecting $N_c$ real-world samples without training. After $N_c$ steps, the training starts.

An additional problem is that the temporal difference error of the value estimate is very high, which is probably caused by the reality gap. Consequently, the critic is changing significantly at the beginning of the real-world training. Training the actor on the varying critic often leads to losses in performance. Therefore, we further delay optimization steps of the actor to $N_a$ steps in total, with $N_a\geq N_c$.
\section{Experimental Setup}
\label{sec:setup}
This section describes the experimental setup for analyzing our proposed architecture and sim-to-real transfer strategies.

\begin{figure}
    \centering
    \vspace{5pt}
    \includegraphics[width=0.8\columnwidth]{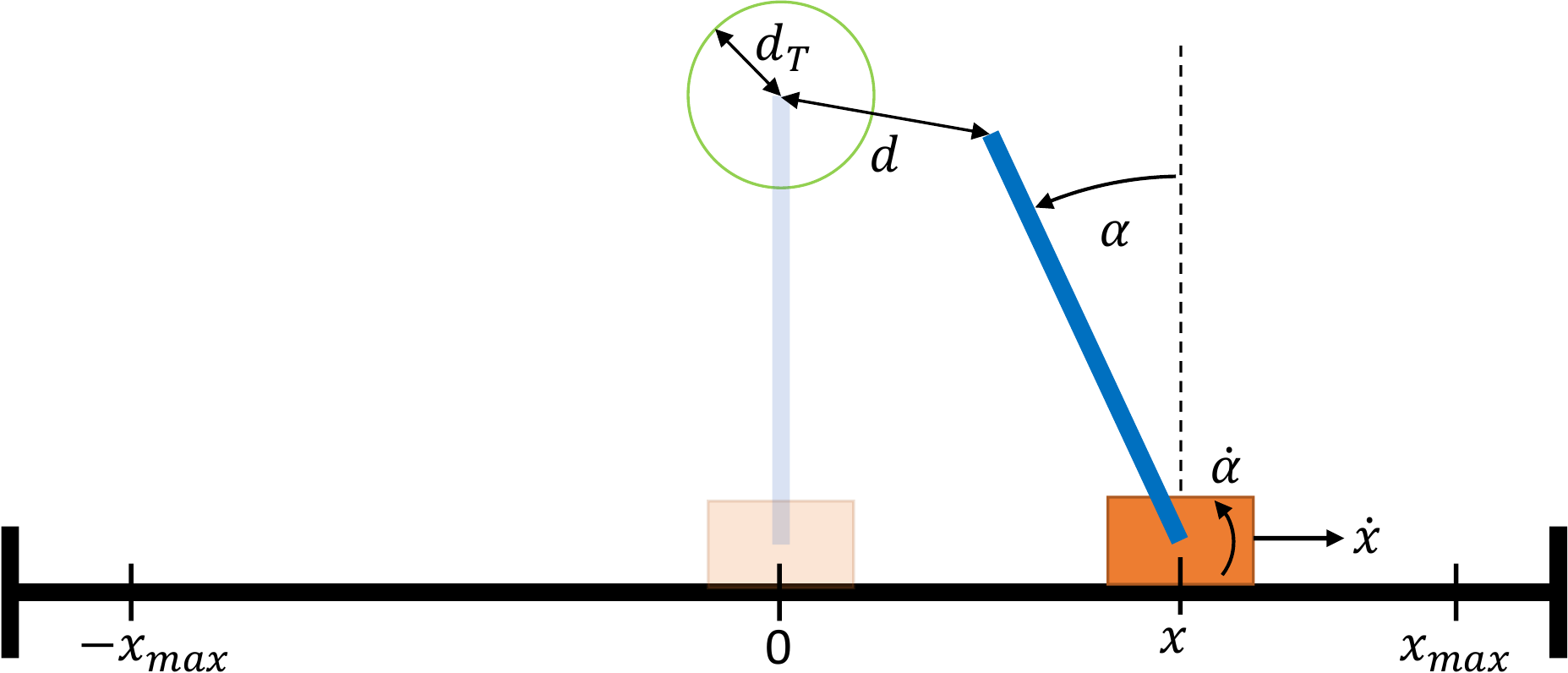}
    \caption{Inverted pendulum system consisting of a cart and a pole with the target configuration faded out in the middle. The Cartesian Euclidean distance from the tip of the pendulum to the target is indicated by $d$.}
    \label{fig:ips}
\end{figure}

\subsection{Inverted Pendulum Control System}
The inverted pendulum control system, schematically shown in Figure \ref{fig:ips}, is a classical benchmark that has been widely studied in the control and real-time domain \cite{seto1999case}. The goal in the inverted pendulum control task is to swing up a pendulum and balance it at the upright position under physical constraints. The inverted pendulum is an inherently unstable non-linear system, making it non-trivial to derive a controller that can swing the pendulum up and balance it. In this work, we formulate the inverted pendulum control problem as a discounted infinite MDP and train a DDPG agent to solve it.

\subsubsection{State, Action and Observation Spaces}
The state of the inverted pendulum consists of four elements $s_t =(x_t,\dot x_t,\alpha_t,\dot\alpha_t)$, with the horizontal cart position $x_t$, its horizontal velocity $\dot x_t$, the angle of the pole with respect to the upright position $\alpha \in (-\pi, \pi] $, and its angular velocity $\dot \alpha_t$. We convert the measured angle $\alpha_t$ into $\sin\alpha_t$ and $\cos\alpha_t$ to simplify the learning process. The control action $a_t \in [-1, 1]$ is the scaling factor of the supply voltage of a DC motor, which drives the cart. Since the delays of the system violate the Markov property, we augment the observation by combing the state observation with the last five actions to make the system observable \cite{ibarz2021train}. The observation space can be expressed as
\begin{equation}\label{observation_space}
o_t= (x_t, \dot x_t, \sin\alpha_t, \cos\alpha_t, \dot\alpha_t, a_{t-5}, \dots, a_{t-1} ),
\end{equation} 
in which $a_{t-5}, \dots, a_{t-1}$ are the previous five actions. We add terminal states to the discounted infinite MDP that stop the process when safety bounds are violated. A state is terminal if
\begin{equation}\label{eq:termination}
    \beta(s_t) =
    \begin{cases}
    1, &\text{if $|x_t| \ge x_{max}$  or  $|\dot\alpha_t| \ge \dot\alpha_{max}$} \\
    0, &\text{otherwise,}
    \end{cases}
\end{equation}
i.e., if either its position is exceeding the bounds of the track or the angular velocity is too high, posing a threat to humans or the system.

\subsubsection{Reward Function}
In the MDP, the agent aims to minimize the Cartesian Euclidean distance $d(s)$ between the tip of the pole and the target position, as shown in Figure \ref{fig:ips}. The target position is given by the position of the tip of the pole when $x=\alpha=0$. We define the reward function as 
\begin{equation}
    R(s_t,a_t,s_{t+1}) = e^{-\delta d(s_t)} - u a_t^2 - v \beta(s_{t+1}).
\end{equation}
The first term is rewarding smaller distances to the target. We chose the exponential function because it limits this term to $(0, 1]$ and has strong gradients when the distance is close to zero. This helps the agent to balance the pendulum exactly on target, avoiding drifts. The second term is penalizing high actions and the third term penalizes safety violations. The parameter $\delta$ stretches the exponential and $u$ and $v$ balance the importance of the action and safety penalties.

\subsection{Evaluation Metrics}

We split the training process into episodes with a maximum length of $T_e$ steps to evaluate the training performance. After every five training episodes, we run an evaluation episode, in which no exploration noise is applied. In these evaluation episodes, we determine if the agent can swing up and balance the pendulum, defined as follows.

The pendulum is defined to be on target if the distance $d(s_t)$ is smaller than a predefined target distance $d_T$. The consecutive steps that the pendulum is on target are called the consecutive on-target-steps and its evolution is defined as
\begin{equation}\label{eq:on_target_step}
    n_{t+1} = 
    \begin{cases}
    n_t+1, & d(s_t) \leq d_T \\
    0, & \text{otherwise.}
    \end{cases}
\end{equation}


We define that the agent can swing up and balance the pendulum if at step $T_e$ the consecutive on-target steps are $n_{T_e} \geq T_g$. 

The training is finished if the agent can reach five consecutive successful evaluation episodes. At this point, we define that the agent converged. The convergence time is the cumulative number of steps in training episodes until the agent converges. During evaluation episodes, no steps are accumulated since no experiences are collected.

\subsection{Simulation Pretraining}

The simulation environment used for pretraining is adapted from the OpenAI-Gym cart-pole environment \cite{OpenAI-gym}, with the parameters set to the values of the physical pendulum. We adapted the original OpenAI-Gym cart-pole environment to use continuous actions and incorporate friction on the cart and pole. To induce different \textit{reality gaps}, we vary a friction factor $k_f$ in simulation, setting the friction at the cart to $k_f$ and at the pole mount to $k_f\times10^{-4}$.

The actor and critic networks in the DDPG algorithm are both implemented as a Multi-Layer Perceptron (MLP) with three fully connected hidden layers of 256, 128, and 64 neurons activated with ReLU. The observations $o_t$ form the input of the actor, and the observations $o_t$ and the action $a_t$ form the input of the critic. The output layers of the actor and critic are a single neuron activated with Tanh and without activation, respectively. 

For simulation training, we use the commonly used OU-noise \cite{uhlenbeck1930theory} with decaying noise magnitude. We reset episodes to a random initial state after either $T_e$ is reached, or if $\beta(s_t)=1$. We further reset the episode when $n_t$ exceeds 100 steps because the experiences collected when the pendulum is on target are very similar and do not improve the learning. These experiences reduce the efficiency of the replay memory as more crucial experiences of the swing-up phase get sampled less likely. The models are pretrained for 1 million steps, with a moving average of the performance determining at which step the best model is saved.

\subsection{Architecture Setup}
The distributed architecture follows a cloud-edge pattern. The cloud device is a workstation with a Xeon Silver-CPU (2.1GHz) and an NVIDIA Quadro RTX 8000 Graphics card. The edge device is a Raspberry Pi 4B board without DNN accelerators. The plant is a linear inverted pendulum built by Quanser \cite{quanser}. The edge device is connected with the plant via USB, through which the sensor readings and control actuation are passed. The cloud and edge are physically connected via Ethernet in a local network. The data communication between the edge and cloud devices is based on TCP packets handled by Redis \cite{redis} hosted on the cloud. 

The double buffered\footnote{When implementing the double-buffering in a multi-threaded Python environment, the global interpreter lock (GIL) needs to be taken into consideration.} actors are deployed on the edge device and interact with the physical system at 30 Hz. The trainer on the cloud optimizes once per experience received from the edge. If the cloud is slower than the edge interactions, a backlog is accumulated and consumed during physical system resets or evaluation episodes where no new experiences are collected. The edge requests new weights from the cloud whenever it finishes applying the previous weights.

As in simulation, we reset the pendulum after $T_e$ steps, or if $\beta(s_t)=1$, or if $n_t$ exceeds 100. The reset is conducted with a simple P-controller which moves the cart to a random position along the track. The angle and angular rate are naturally randomized. Additionally, every 10,000 steps, we calibrate the angular encoder by letting the pendulum settle and reset the angle to $\pi$ radians. This is to prevent angle drift over long training sessions. A list of the parameters can be found in Table \ref{table:parameters}.

\begin{table}
\vspace{5pt}
\aboverulesep=0ex 
\belowrulesep=0ex 
\center
\small
\begin{tabular}{c|c||c|c}
\toprule[1.5pt]
\rule{0pt}{1.1EM}%
Parameter & Value & Parameter & Value \\
\midrule
\rule{0pt}{1.1EM}%
$x_{\text{max}}$ & 0.34 m &$\delta$ & 5 \\
$\dot\alpha_{\text{max}}$ & 20 rad/s &$u$ & 0.1 \\
$d_T$ & 0.05 m & $v$ & 20  \\
 $k_f$ & 10 & $N_c$ & 3500 \\
 $T_e$ & 1000 &  $N_a$ & 5000\\
 $T_g$ & 750 & $B$ & 128 \\
\bottomrule[1.5pt]
\end{tabular}
\caption{Default parameters for the experiments.}
\label{table:parameters}
\end{table}
\section{Results}
\label{sec:results}
In this work, we are interested in the following three questions:
\begin{enumerate}
    \item How are different sizes of the \textit{reality gap} affecting training time?
    \item How do various combinations of optimization delays influence training performance on the physical system?
    \item Is the communication latency between the cloud and edge affecting training performance?
\end{enumerate}

We assume that these experimental variables are orthogonal to each other. Therefore, we vary these parameters independently and compare the convergence time of corresponding real-world training instances. A comparison between pretraining in simulation and directly training in the real world was not possible. When training from scratch, the randomly initialized agent was too aggressive, repeatedly causing damage to the physical system. 

For all the experiments, if not indicated otherwise, the parameters used are listed in Table \ref{table:parameters}. We conducted five real-world training instances for each configuration, on average trained for around one hour each.

\subsection{Reality Gap}

\begin{figure}
    \centering
    \vspace{5pt}
    \includegraphics[width=0.8\columnwidth]{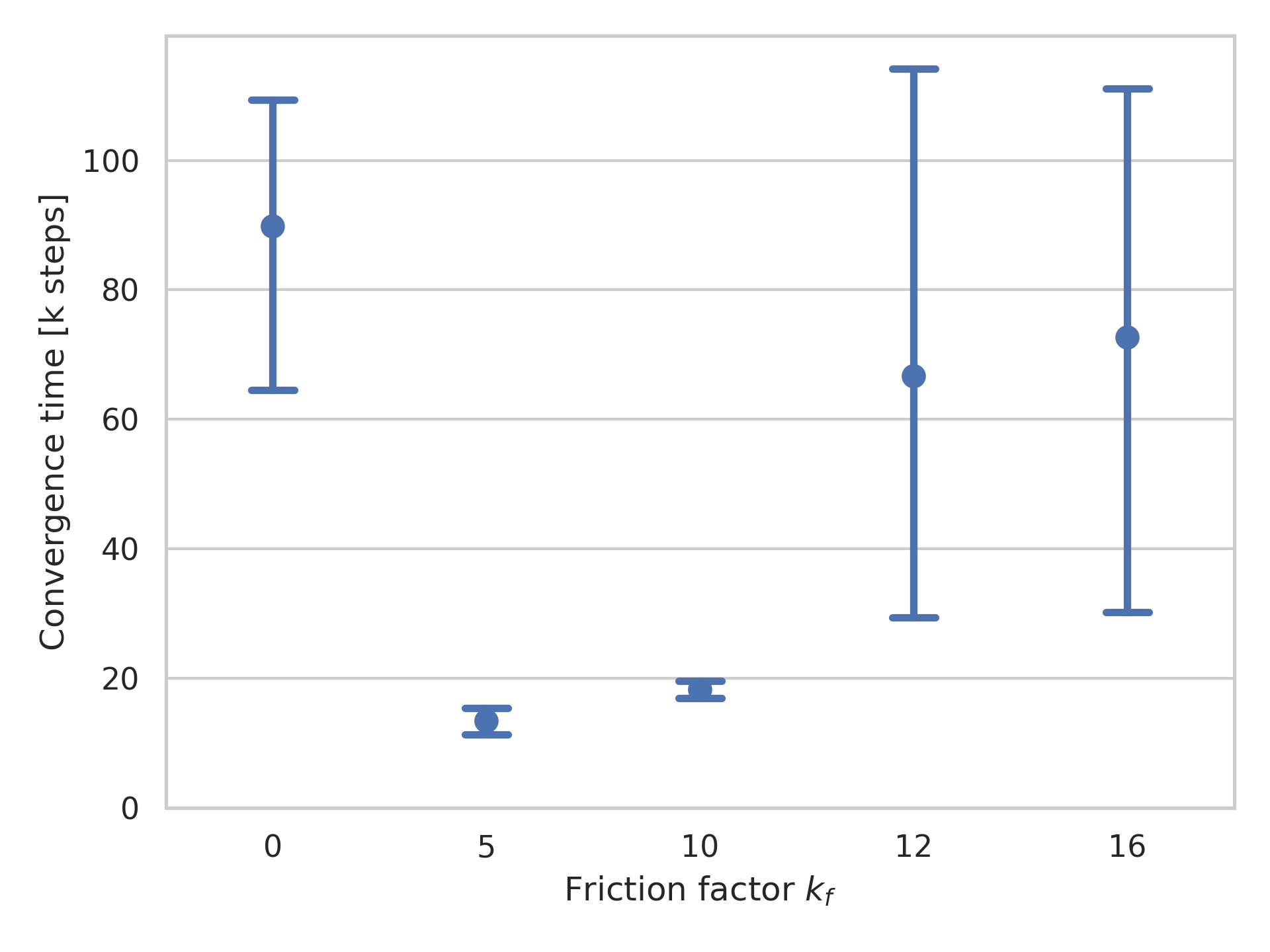}
    \caption{Convergence time comparison of real-world training instances with different models pretrained with different friction factors.}
    \label{fig:friction_vars}
\end{figure}

\begin{figure*}
    \centering
    \vspace{5pt}
    \includegraphics[width=\textwidth]{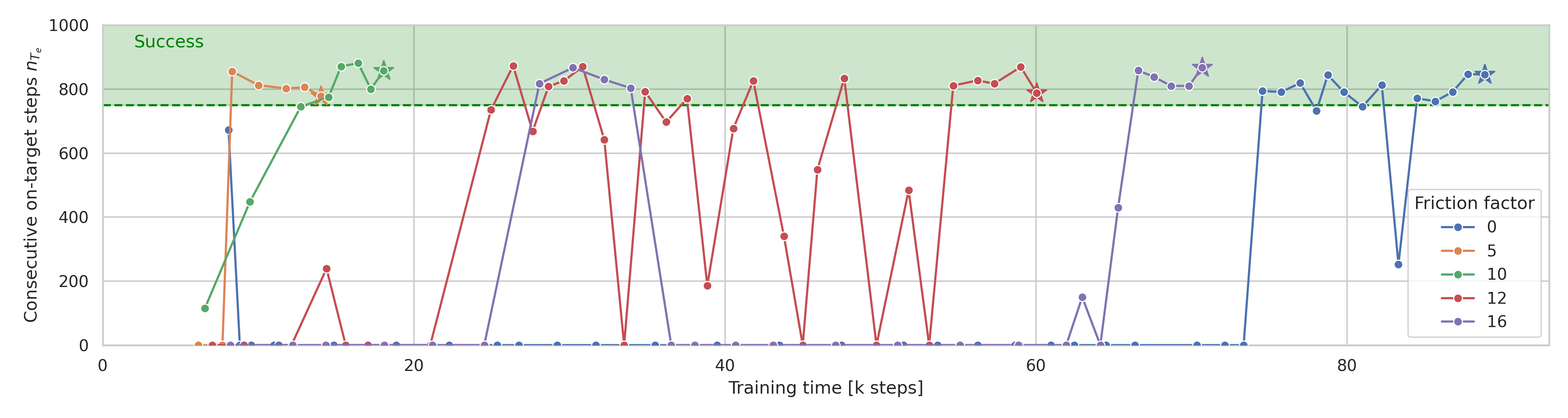}
    \caption{Time-series plot showing the learning progress of five models pretrained with different friction factors, indicated by the color. Each data point corresponds to an evaluation episode with the horizontal axis' cumulative training steps and the vertical axis' consecutive on-target steps. Data points with more than 750 consecutive on-target steps are classified as a success. The fifth success data point in a row is highlighted with a star, indicating that the training converged.}
    \label{fig:training_plot}
\end{figure*}

To study the influence of the \textit{reality gap}, we vary the friction properties of the system, as they are hard to model and can thus be an essential contributor to the \textit{reality gap} \cite{ChristianoSMSBT16}. We vary the friction factor $k_f$ in simulation, setting its value to one of $\{0,5,10,12,16\}$. For $k_f=0$, the system is frictionless, and $k_f=16$ results in the maximum friction under which the agent can learn to swing up the pendulum. For $k_f>16$, the agent does not learn repeatably, which we attribute to the high energy losses. 

The convergence times for each configuration are depicted in Figure \ref{fig:friction_vars}. The results show that the convergence time is lowest for $k_f=5$, closely followed by $k_f=10$. When training with $k_f=0$, the pendulum does not reach the target at the beginning of the training where it overshoots the target with $k_f > 10$. This is expected as, compared with the simulation, the friction decelerates the pendulum stronger or weaker, respectively.

Taking a closer look at the training process, Figure \ref{fig:training_plot} shows the training time-series for the training instances that resulted in the median convergence time of each configuration. It can be seen that the training instances for $k_f=5$ and $k_f=10$ converged very fast after only a few evaluation episodes. The other configurations required significantly more training time. The frictionless pretrained agent needs to learn to swing up but is already sufficiently good at balancing, since in the frictionless simulation, the swing-up is easier and the balancing is harder. This can be observed in the training time-series, as the $k_f=0$ agent requires many training steps to succeed once, corresponding to learning to swing up, but then converges quickly as it can already balance well. The pretrained agents with $k_f=12$ and $k_f=16$ succeed earlier, indicating that they can swing up but require more training steps to learn how to stabilize the pendulum consistently. This might have been exacerbated by our early resetting policy, which was supposed to balance the replay memory by reducing the experiences with a stabilized pendulum. This strategy could have inadvertently favored pretrained agents that were already good at balancing and needed to learn to swing up and disadvantaged agents that needed to learn to stabilize the pendulum.

Generally, it can be observed that using the proposed distributed cloud-edge architecture with the transfer strategies, pretrained models with different \textit{reality gaps} can learn to swing up and balance the inverted pendulum in the real world repeatably and consistently. For the following experiments, we selected the pretrained model with $k_f=10$ as it cannot succeed immediately, requiring real-world training. It further has low variance, making the following comparisons easier.

\subsection{Optimization Delay}

\begin{figure}
    \vspace{-10pt}
    \centering
    \includegraphics[width=0.8\columnwidth]{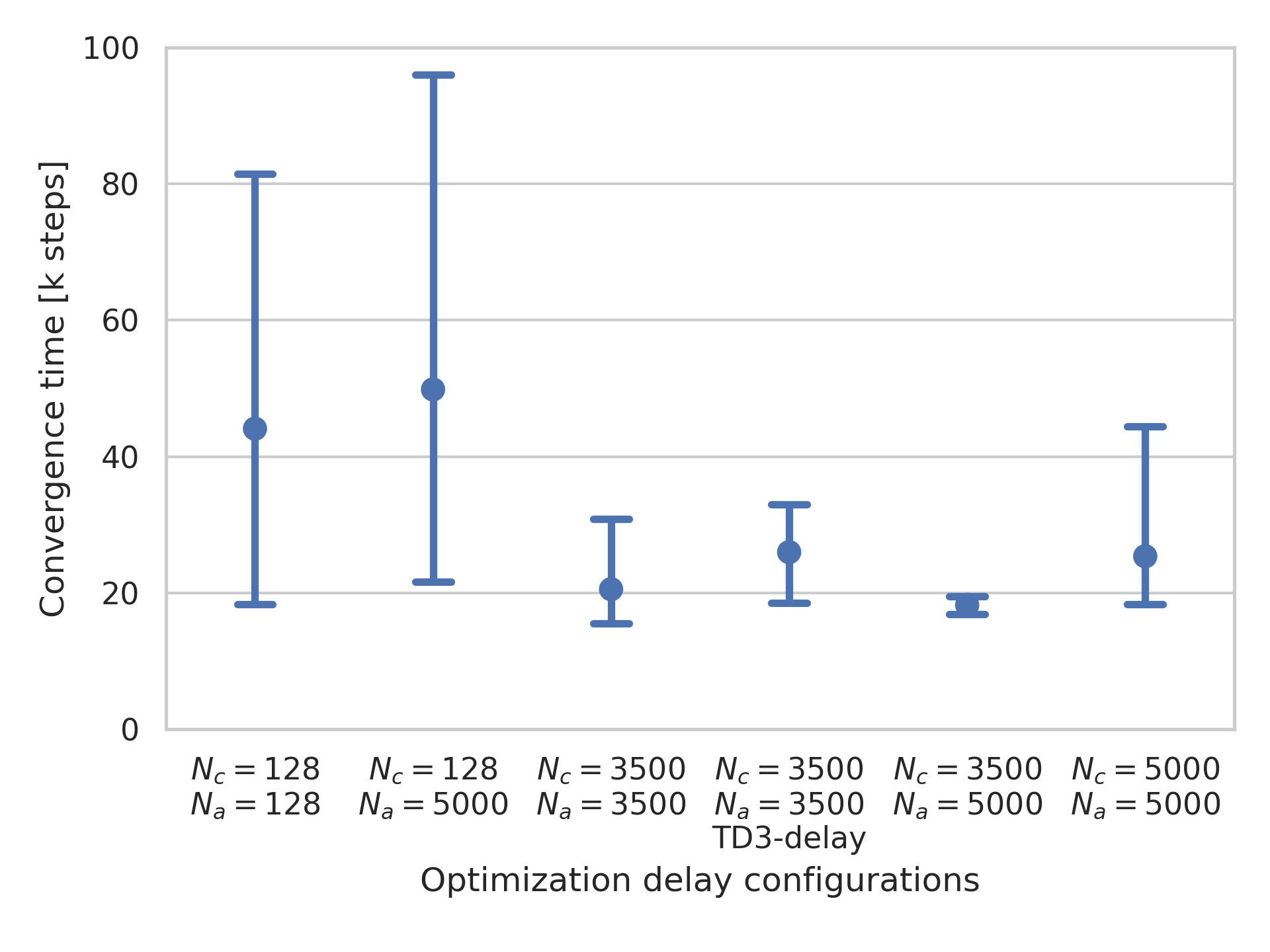}
    \caption{Convergence time of different optimization delay configurations including one TD3-like actor delay implementation.}
    \vspace{-5pt}
    \label{fig:freezing_vars}
\end{figure}

To study the effect of the optimization delays $N_c$ and $N_a$ on the convergence time we compare different delay combinations out of $\{128,3500,5000\}$ for $N_c$ and $N_a$ in Figure \ref{fig:freezing_vars}. The minimal delay is the batch size $B=128$. Since the actor is trained on the critic, we only investigate configurations with $N_a \geq N_c$. Additionally, as the actor optimization delay has similarities with the delayed training in TD3, we tested one configuration with the actor delay implementation of TD3.

The first conclusion from the data is that an $N_c=B$ is not sufficient since the replay memory cannot decorrelate experience samples, leading to unpredictable learning behavior of the critic. The bad learning performance of the critic cannot be recovered by delaying the actor optimization as seen for $N_a=5000$. Additionally, a high $N_c=5000$, appears to be less beneficial than $N_c=3500$. However, this effect is marginal, possibly caused by statistical errors due to few samples. The second observation is that the additional actor delay $N_a > N_c$ seems to give a slight advantage since the combination $N_c=3500$ and $N_a=5000$ performed the best. The introduced TD3-delay did not improve training performance.

This experiment concludes that the prefilling of the replay memory by delaying the critic optimization is essential for stable learning performance. The additional actor optimization delay may slightly improve the learning performance, but the effect is marginal.

\subsection{Cloud-Edge Communication Latency}

\begin{figure}
    \vspace{-10pt}
    \centering
    \includegraphics[width=0.8\columnwidth]{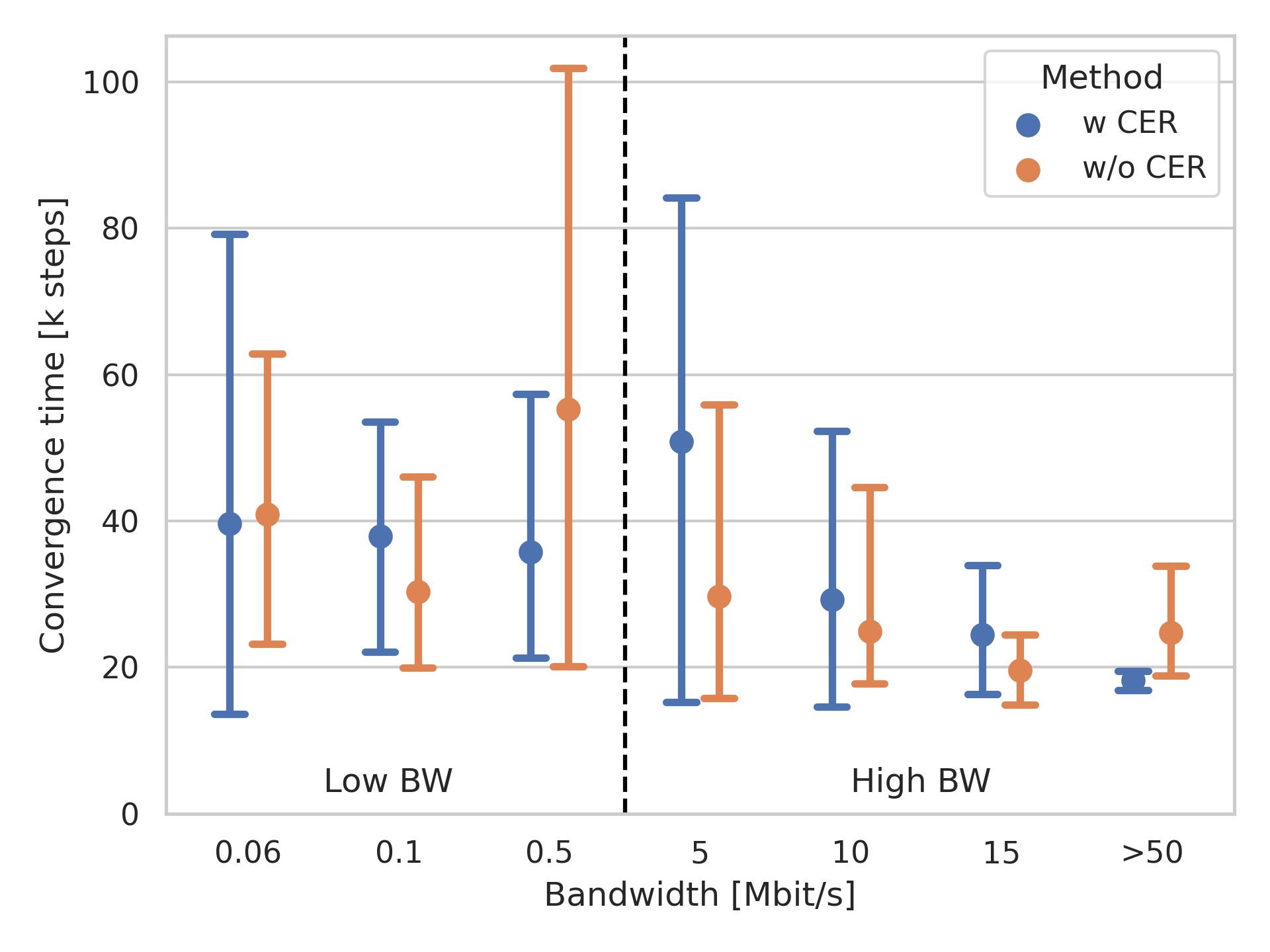}
    \caption{Convergence time comparison of training instances with different bandwidth settings classified in low and high, trained with and without CER.}
    \vspace{-5pt}
    \label{fig:bandwidth_vars}
\end{figure}

With the last experiments, we aim to analyze the sensitivity of the proposed architecture to the data bandwidth between the cloud and the edge. A constrained bandwidth reduces the actor update frequency, potentially changing learning performance. To this end, we artificially constrained the bandwidth by delaying actor packets accordingly. For the experiments, we selected bandwidths from $\{0.06, 0.1, 0.5, 5, 10, 15, \footnotesize{>}50\}$ Mbit/s which correspond to an actor update roughly every $\{681,  408,  81, 8, 4, 2, 1\}$ interaction steps. The lowest value, 0.06 Mbit/s, is the minimum bandwidth needed from the edge to the cloud to send state-action pairs continuously. The networked Ethernet setup in the previous experiments corresponds to $\footnotesize{>}50$ Mbit/s. The bandwidth values can be split into two groups, a low-bandwidth region ($\{0.06, 0.1, 0.5\}$) and a high-bandwidth region ($\{5, 10, 15, \footnotesize{>}50\}$). Since we suspected that combined experience replay (CER) should be affected by the bandwidth, we conducted all experiments with and without the usage of CER.

Figure \ref{fig:bandwidth_vars} shows a clear trend in the high-bandwidth region that for lower bandwidths, average convergence time and variance increase. The convergence time average and variance do not increase further in the low-bandwidth region. This trend is similar with and without CER. A noteworthy observation is that CER only improves performance without a bandwidth constraint. For most other bandwidth constraints, CER decreases performance. We hypothesize that the on-policy behavior induced by CER biases the training process if the actor does not change every step. This hypothesis could explain the performance reduction for the other configurations.

This experiment concludes that the proposed architecture and transfer strategies perform best under high bandwidths from the cloud to the edge. However, agents can be trained within reasonable time even under highly constrained bandwidths. A side conclusion is that CER is only beneficial if the actor is updated on every interaction step. Otherwise, it is even detrimental, which is a significant limitation for the usage of CER.
\section{Conclusions and Future Work}
\label{sec:conclusion}
Training deep reinforcement learning agents on physical systems is challenging due to expensive data collection. Sim-to-real approaches train agents in simulations and directly deploy them to the physical system, suffering from performance losses induced by the \textit{reality gap}. Sim-to-real approaches with continued training on the physical system may be the best compromise of simulation and real-world training, which is why we adopt this strategy for this work.

In this work, we propose a distributed cloud-edge architecture to address the problem of training reinforcement learning agents on computationally constrained physical systems. We further propose sim-to-real transfer strategies with delayed optimization of the neural networks and introduce a one-step delay in simulation to force the agent to learn from its observation augmentation.

We evaluate the proposed methods with a linear inverted pendulum case study. We vary the \textit{reality gap}, optimization delays, and cloud-edge bandwidth to analyze the performance of the proposed architecture. As expected, higher \textit{reality gaps} lead to longer training time. However, it is still possible to train agents with high \textit{reality gaps} repeatably using the architecture. After the sim-to-real transfer, we identified a crucial replay memory prefill that stabilizes the learning process. The training is faster with higher bandwidths, but the architecture still allows training with very constrained bandwidths. An additional conclusion is that CER \cite{zhang2017deeper} is only beneficial if the actor can be updated every interaction step, and even detrimental if the updates happen less frequently. A final observation is that with the cloud-edge architecture, the training process on the cloud is robust to crashes of the plant or edge.

Training data from physical systems often contains noise and disturbances. These faulty experiences can delay the entire training process. We will filter faulty experiences in future work to stabilize the training process further. Moreover, to reduce the sensitivity to the bandwidth, we will look into the usage of residual networks, where only smaller residual networks need to be trained and sent to the edge. Additionally, we will expand the architecture to other applications such as UAVs and use other state-of-the-art DRL algorithms.





\section*{ACKNOWLEDGMENT}
Marco Caccamo was supported by an Alexander von Humboldt Professorship endowed by the German Federal Ministry of Education and Research. 

The authors would like to thank Daniele Bernardini, Andrea Bastoni, Alexander Züpke and Andres Rodrigo Zapata Rodriguez for helpful discussions.

\balance
\bibliographystyle{ieeetr}
\bibliography{ref}

\end{document}